\documentclass{article}

\usepackage[preprint]{neurips_2025} 

\usepackage{amsmath,amsfonts,amssymb,mathtools}
\usepackage{graphicx}
\usepackage{wrapfig}
\usepackage{booktabs}
\usepackage{hyperref} 
\usepackage{enumitem} 
\usepackage{natbib}
\usepackage{xcolor}
\usepackage{float}

\defcitealias{artificialanalysis2025}{AA, 2025} 

\allowdisplaybreaks

\title{Compute-Accuracy Pareto Frontiers for Open-Source Reasoning Large Language Models}
\author{
  Ákos Prucs\textsuperscript{1,}\thanks{Correspondence to \texttt{akos.prucs@egroup.hu}} \quad
  Márton Csutora\textsuperscript{1,2} \quad
  Mátyás Antal\textsuperscript{1,2} \quad
  Márk Marosi\textsuperscript{1,2} 
  \\[1em] 
  \textsuperscript{1}E-Group Research
  \textsuperscript{2}Budapest University of Technology and Economics \\
}

\begin{document}

\maketitle

\begin{abstract}
Large Language Models (LLMs) are demonstrating rapid improvements on complex reasoning benchmarks, particularly when allowed to utilize intermediate reasoning steps before converging on a final solution. However, current literature often overlooks the significant computational burden associated with generating long reasoning sequences. For industrial applications, model selection depends not only on raw accuracy but also on resource constraints and inference costs. 
In this work, we conduct a test-time-compute aware evaluation of both contemporary and older open-source LLMs, mapping their Pareto frontiers across math- and reasoning-intensive benchmarks. Our findings identify the Mixture of Experts (MoE) architecture as a strong candidate to balance performance and efficiency in our evaluation setting. Furthermore, we trace the trajectory of Pareto efficiency over time to derive an emergent trend regarding accuracy gain per unit of compute. Finally, we demonstrate that there is a saturation point for inference-time compute. Beyond a certain threshold, accuracy gains diminish, indicating that while extended reasoning capabilities are beneficial, they cannot overcome intrinsic model limitations regarding specific complexities.

\end{abstract}

\section{Introduction}

The rapid evolution of Large Language Models (LLMs) has been initially driven by empirical scaling laws, which established power-law relationships linking model performance to the scale of parameters~\citep{kaplan2020scaling, brown2020gpt3}, training data~\citep{hoffmann2022training}, and pre-training compute~\citep{kaplan2020scaling, sutton2019bitter}. While the dominant paradigm focused on training larger dense models, parallel architectural advancements, specifically efficient attention mechanisms, have dramatically expanded context windows~\citep{dao2022flashattention, chen2023longloRA}. This expanded capacity has been pivotal for effectively removing the ceiling on generation length and allowing models to utilize the context as a working memory for extensive Chain-of-Thought (CoT) prompting~\citep{wei2022chain, nye2021show}. This shift suggests that the frontier of performance is no longer defined solely by pre-training scale, but increasingly by the scaling of test-time compute~\citep{snell2024scaling, brown2024language}. As the first generation of reasoning-specialized models demonstrates~\citep{openai2024o1, jiang2024deepseekr1, geminiteam2024gemini}, allocating additional computational resources to generate thousands of intermediate reasoning tokens can yield significant accuracy gains on complex benchmarks, necessitating a rigorous re-evaluation of efficiency that accounts for this new inference cost.

However, the prevailing focus on maximizing leaderboard performance obscures the practical trade-offs inherent in test-time scaling. Model publishers typically highlight mere accuracy metrics, implicitly assuming an unconstrained computational resource at inference time. This accuracy-centric view is misleading for industrial applications, where deployment decisions are governed by strict latency requirements, hardware limitations, and unit economics. While standard evaluations effectively measure a model’s maximal capability, they often overlook the computational cost of extended CoT reasoning, thereby failing to capture the critical dimension of inference efficiency. To make informed deployment choices, it is essential to expose the specific computational expenditure required to achieve a given level of accuracy; by mapping this relationship, we construct the empirical Pareto frontier of reasoning accuracy against floating-point operations. This framework renders the identification of optimal models accessible, enabling practitioners to navigate the landscape of test-time scaling and pinpoint architectures that maximize performance within strict compute budgets.

\begin{figure*}[!t]
    \centering
    \includegraphics[width=\textwidth]{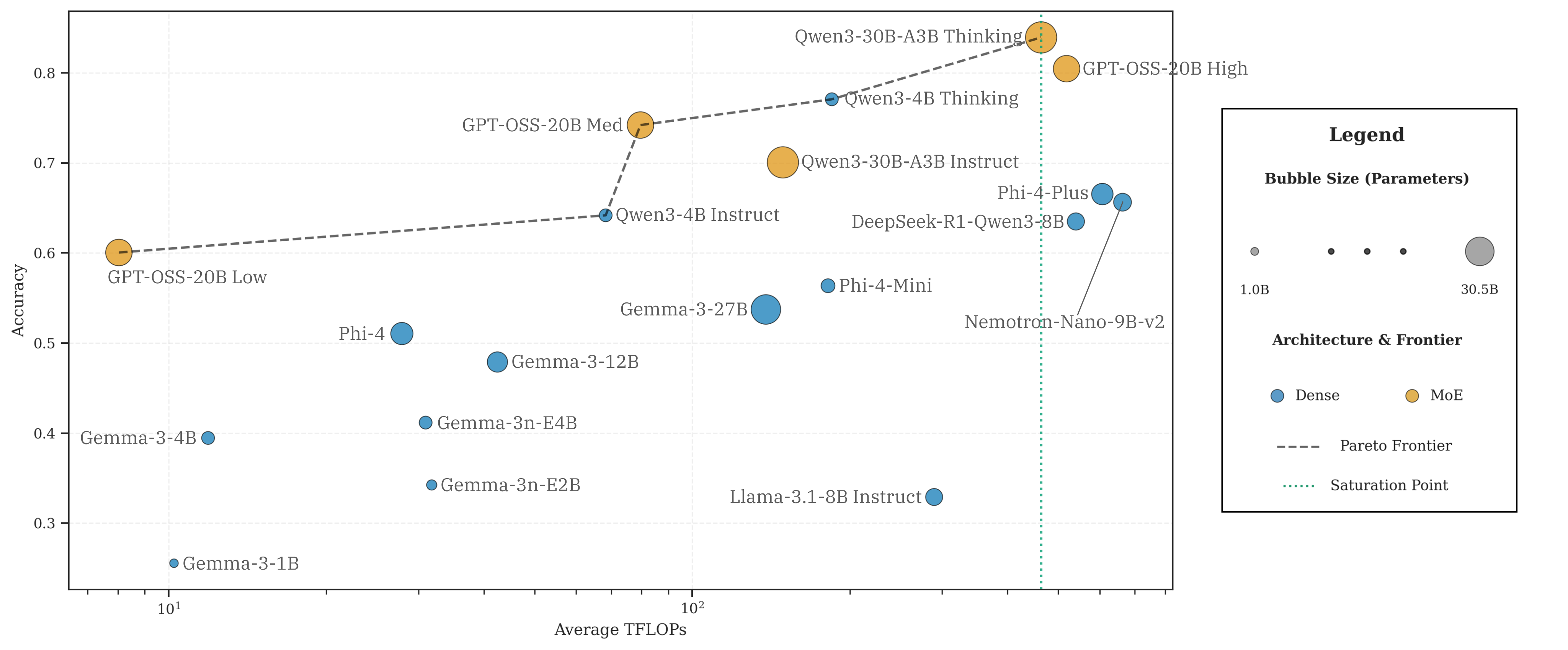}
    \caption{\textbf{The Reasoning Pareto Frontier.} 
    Each point represents the efficiency of a model averaged across the five benchmarks. The $x$-axis shows the average KV-aware estimation of FLOPs per query (log scale), and the $y$-axis shows task accuracy. 
    \textbf{Color} indicates model type (dense in blue, mixture of experts in yellow), and \textbf{marker size} is proportional to the total model parameter count.}
    \label{fig:pareto-main}
\end{figure*}

\paragraph{Contributions.}
In this work, we perform a compute-aware, single-pass, CoT evaluation of contemporary and traditional models, focusing on instruction and reasoning fine-tuned (FT) variants. Each model is evaluated on math and reasoning benchmarks of variable difficulty levels.  

Our contributions are the following:
\begin{enumerate}[label=\textbf{(\roman*)}, leftmargin=*, parsep=3pt]
\item mapping the empirical Pareto frontier between task accuracy and inference compute for all contemporary models across five benchmarks, providing an overview of model performance in a compute aware system;
\item demonstrating an architectural advantage for Mixture-of-Experts (MoE) models, which consistently lead the Pareto frontier by achieving higher accuracy than dense models at comparable FLOP budgets;
\item tracing the trajectory of Pareto efficiency over time to derive an emergent trend, providing a statistical basis for predicting the performance efficiency of upcoming models;
\item identifying a task-dependent Pareto knee beyond which additional compute provides negligible accuracy gain, with easier benchmarks plateauing at lower inference computes compared to harder competition-style and graduate-level problems;

\item finding that smaller models can nonetheless occupy Pareto-optimal points by using substantially extended reasoning traces to compensate for reduced parameter counts;
\item highlighting an asymmetry in compute usage, whereby incorrect answers systematically consume more compute than correct ones, indicating that models expend significantly more computational resources attempting to solve problems outside their capabilities
\end{enumerate}


\section{Related Work}

The evolution of LLM reasoning capabilities has been driven by the development of effective inference-time strategies ~\citep{wei2022cot,yao2023tree,wang2022selfconsistency,zhou2023leasttomost} alongside the optimization of specialized models through reinforcement learning ~\citep{openai2024o1,jiang2024deepseekr1,cobbe2021gsm8k,zelikman2022star,lightman2023let}. A unifying pattern across this spectrum is the utilization of \emph{test-time compute scaling}, where extended reasoning traces yield higher accuracy ~\citep{snell2024testtime,brown2024o1scaling}. However, this performance gain incurs steep infrastructural costs, particularly regarding inference latency and throughput, that persist despite recent architectural and low-precision optimizations ~\citep{dao2022flashattention,kwon2023vllm,dettmers2022llmint8,frantar2022gptq,lin2023awq}. While prior work has applied Pareto analysis to training efficiency ~\citep{hoffmann2022chinchilla}, recent studies have begun to characterize inference-time trade-offs regarding energy consumption, verifier-free scaling, and cost-accuracy frontiers ~\citep{jin2025energy, xiao2025densing, kim2025parameter}.  We extend this emerging landscape by specifically mapping the FLOPs-accuracy Pareto frontiers of open-source models, offering a multi-dimensional framework that explicitly weighs reasoning efficacy against computational cost.

\section{Methodology}
This section details the framework for our single-pass, CoT evaluation. We first outline the selected benchmarks and model landscape, followed by the prompting, decoding, and answer extraction strategies. Finally, we derive the FLOPs estimation formulas used to quantify inference efficiency.

\subsection{Benchmarks and Models}

We evaluate 19 open-source state-of-the-art (SOTA) models across a diverse suite of reasoning tasks, ranging from grade-school arithmetic (\textsc{gsm8k}~\cite{cobbe2021training}) and competition-level mathematics (\textsc{AIME25}~\cite{maa2025aime}, \textsc{HMMT-FEB25}~\cite{hmmt2025feb}, \textsc{MATH500}~\cite{lightman2023lets}) to graduate-level scientific problems (\textsc{GPQA-Diamond})~\cite{rein2023gpqa}.

Our model pool includes both dense~\cite{dubey2024llama, abdin2024phi, deepseek2025deepseek, team2025gemma, qwen2025qwen3} and MoE architectures~\cite{qwen2025qwen3, openai2025gptoss}, spanning parameter scales from small models at one billion parameters to larger models up to tens of billions of parameters. Additionally, we include Nemotron-Nano-V2~\cite{nvidia2025nemotron} in our evaluation to represent emerging Mamba-Transformer hybrid architectures. Models are instantiated using their default inference configurations and official system prompt templates to ensure faithful reproduction of their intended behavior. Where models offer adjustable reasoning profiles, we treat each configuration as a distinct entry; for instance, GPT-OSS~\cite{openai2025gptoss} is evaluated separately across its low, medium, and high modes.

\subsection{Prompting and Decoding}

Our evaluation methodology aligns with the intelligence benchmarking protocols established by Artificial Analysis \citepalias{artificialanalysis2025}. As they provide a unified structure for intelligence assessment, we explicitly adopt their recommended prompting templates, temperature parameters, and judge evaluation prompts across all benchmarks. This ensures a consistent testing environment where performance differences can be attributed to intrinsic model capability rather than variance in prompt engineering.

Being consistent with the AA guidelines, we distinguish between sampling configurations based on model type:
\begin{itemize}
    \item \textbf{Reasoning-tuned models:} We use a temperature of \(T=0.6\), facilitating the exploration required for complex CoT paths.
    \item \textbf{Standard instruction models:} We employ greedy decoding (\(T=0\)) to maximize determinism.
\end{itemize}

Beyond temperature, we prioritize the model-default values for \texttt{top\_k} and \texttt{top\_p} as specified in each model's official documentation. Additionally, we universally apply a small frequency penalty of \(0.05\), which we found empirically effective in preventing the repetitive loops that occasionally occur during long-context generation. Regarding the generation budget, we evaluate a single completion per query, setting the maximum token limit to the full extent of each model's architectural context window. This ensures that reasoning traces are terminated exclusively by the model's natural stop criteria rather than arbitrary length constraints.

\subsection{Evaluation protocol}

To ensure a consistent assessment of reasoning capabilities and minimize false negatives in answer verification, we implement a hierarchical, two-pass evaluation system. This approach combines deterministic parsing with semantic verification to handle the variability of CoT outputs. 

\paragraph{Input Standardization and Regex Extraction.}
To facilitate automated evaluation, models are prompted to adhere to a unified output format. Specifically, models are instructed to present their final conclusion either enclosed within a LaTeX \verb|\boxed{...}| element or immediately following a distinctive \texttt{Answer:} token sequence. Our primary extraction pass utilizes a deterministic regular expression parser targeting these patterns. This method prioritizes precision and is computationally efficient for well-structured outputs.

\paragraph{Fallback LLM Judge.}
In instances where the regex parser fails to extract a candidate answer, often due to deviations in formatting or verbose conversational closures, we employ a robust, lightweight LLM judge as a fallback mechanism. In this case, we utilize \texttt{gpt-5.1-mini} as this model has a good balance in cost-efficiency and semantic reasoning capabilities. The judge is provided with a template containing the problem statement, the ground-truth (gold) answer, and the model's generated output.

\paragraph{Judge Efficiency and Robustness.}
To optimize the evaluation budget without compromising accuracy, we leverage the structural characteristic of CoT traces where the conclusion uniformly appears at the end of the sequence. Consequently, we truncate the model's output passed to the judge to the final 20 tokens. The judge is tasked with a binary classification: verifying if the gold answer is semantically present within this window.
\begin{itemize}
    \item \textbf{Arithmetic Robustness:} For mathematical tasks, the judge is explicitly instructed to reconcile representational differences, accepting valid variations (e.g., differing floating-point formats, the presence of currency symbols like \texttt{\$}, or answers still wrapped in Latex commands) that strict string matching would reject.
    \item \textbf{Logical Entailment:} For multiple-choice and logical tasks, the judge evaluates whether the model's final tokens entail the correct option, ensuring that the essence of the answer is captured even if the exact string match is absent.
\end{itemize}

\subsection{FLOPs Estimation}
\label{sec:flops-estimation}

Estimating the computational cost of transformer inference is essential for understanding efficiency trade-offs across model families. A widely used approximation treats inference FLOPs as $2N$ per token, where $N$ is the number of non-embedding parameters~\citep{kaplan2020scaling,hoffmann2022chinchilla}. This rule of thumb captures the dominant cost of matrix multiplications, where each parameter contributes one multiply-accumulate (2 FLOPs) per token, but neglects the quadratic cost of attention and assumes architectural homogeneity. Modern transformers employ grouped-query attention (GQA), gated feed-forward networks (SwiGLU), and mixture-of-experts (MoE), each affecting compute in distinct ways. We derive a component-aware estimator that accounts for these variations.

\paragraph{Attention Projections.}
For a transformer with hidden dimension $d$, attention heads $n_h$, key-value heads $n_{kv}$, and head dimension $d_h = d/n_h$, the projection costs per token are:
\begin{equation}
F_{\text{proj}} = \underbrace{2d \cdot n_h d_h}_{\text{Q}} + \underbrace{2 \cdot 2d \cdot n_{kv} d_h}_{\text{K, V}} + \underbrace{2 \cdot n_h d_h \cdot d}_{\text{O}} = 4d^2\left(1 + \frac{n_{kv}}{n_h}\right).
\end{equation}
For multi-head attention (MHA) where $n_{kv} = n_h$, this yields $8d^2$. Grouped-query attention with ratio $r = n_h / n_{kv}$ reduces the K and V projections:
\begin{equation}
F_{\text{proj}}^{\text{GQA}} = 4d^2\left(1 + \frac{1}{r}\right).
\end{equation}
GQA-8 (as in Llama-3-70B) costs $4.5d^2$, marking a 44\% reduction relative to MHA.

\paragraph{Feed-Forward Network.}
With intermediate dimension $d_{\text{ff}}$, modern architectures employ either gated or standard FFN:
\begin{equation}
F_{\text{FFN}} = 
\begin{cases}
6 \, d \, d_{\text{ff}} & \text{(gated: gate, up, down)} \\
4 \, d \, d_{\text{ff}} & \text{(standard: up, down)}
\end{cases}
\end{equation}
The gated variant (SwiGLU), used in Llama, Mistral, Qwen, and Gemma, employs three weight matrices instead of two. To maintain approximately equal parameter count and FLOPs as a standard FFN with $d_{\text{ff}} = 4d$, SwiGLU models typically set $d_{\text{ff}} = \frac{8}{3}d \approx 2.67d$, yielding $6d \cdot \frac{8}{3}d = 16d^2$ versus $4d \cdot 4d = 16d^2$ for standard FFN. In practice, $d_{\text{ff}}$ is rounded to a multiple of 256 for hardware efficiency, resulting in values like $d_{\text{ff}} = 2.75d$ (Llama-2) or $d_{\text{ff}} = 3.5d$ (Llama-3, Mistral).

\paragraph{Mixture-of-Experts.}
For MoE models with $E$ total experts and $k$ active per token:
\begin{equation}
F_{\text{FFN}}^{\text{MoE}} = k \cdot F_{\text{FFN}} + 2dE,
\end{equation}
where $2dE$ accounts for the router. Mixtral-8$\times$7B activates 2 of 8 experts, using only 25\% of FFN parameters per token.

\paragraph{Attention Computation.}
The quadratic attention cost for sequence length $s$ comprises the query-key dot product and value aggregation:
\begin{equation}
F_{\text{attn}}(s) = 4ds^2.
\end{equation}

\paragraph{Mamba Layers.}
Mamba-2 replaces attention with a state-space module whose computation scales
linearly in sequence length. For hidden dimension $d$, expansion factor
$d_{\text{in}} = E d$, state dimension $N$, and convolution kernel size $K$, the
per-token FLOPs of a Mamba layer are approximated as
\begin{equation}
F_{\text{mamba}}
= 
\underbrace{6 d \, d_{\text{in}}}_{\text{input/output projections}}
\;+\;
\underbrace{2 d_{\text{in}} K}_{\text{depthwise convolution}}
\;+\;
\underbrace{2 d_{\text{in}} N}_{\text{SSM/scan}}.
\end{equation}

\paragraph{Total FLOPs.}
For prefill of $P$ tokens and autoregressive generation of $G$ tokens,
let $S = P + G$ denote the total number of processed tokens.  Attention
layers $L_{\text{attn}}$ incur both linear and quadratic costs, whereas Mamba layers $L_{\text{ssm}}$ are
strictly linear in $S$.  The total FLOPs are
\begin{equation}
\begin{split}
F(P,G)
&=
L_{\text{attn}} S \bigl(F_{\text{proj}} + F_{\text{FFN}}\bigr)
+
L_{\text{ssm}} S \bigl(F_{\text{mamba}} + F_{\text{FFN}}\bigr) \\[4pt]
&\quad+\;
4 d\, L_{\text{attn}}
\left(
P^{2} + GP + \frac{G(G+1)}{2}
\right) \\[4pt]
&\quad+\;
2 d V S
\end{split}
\label{eq:flops-total}
\end{equation}

\section{Results}
\label{sec:results}
We present an empirical analysis of 19 state-of-the-art models under single-pass, full Chain-of-Thought (CoT) decoding. Our findings characterize the relationship between computational expenditure and reasoning fidelity, highlighting architectural efficiencies and critical saturation points in test-time scaling.

\subsection{Capabilities and Constraints of Test-Time Compute}
\label{sec:main-pareto}

\begin{figure*}[ht]
    \centering
    \includegraphics[width=\textwidth]{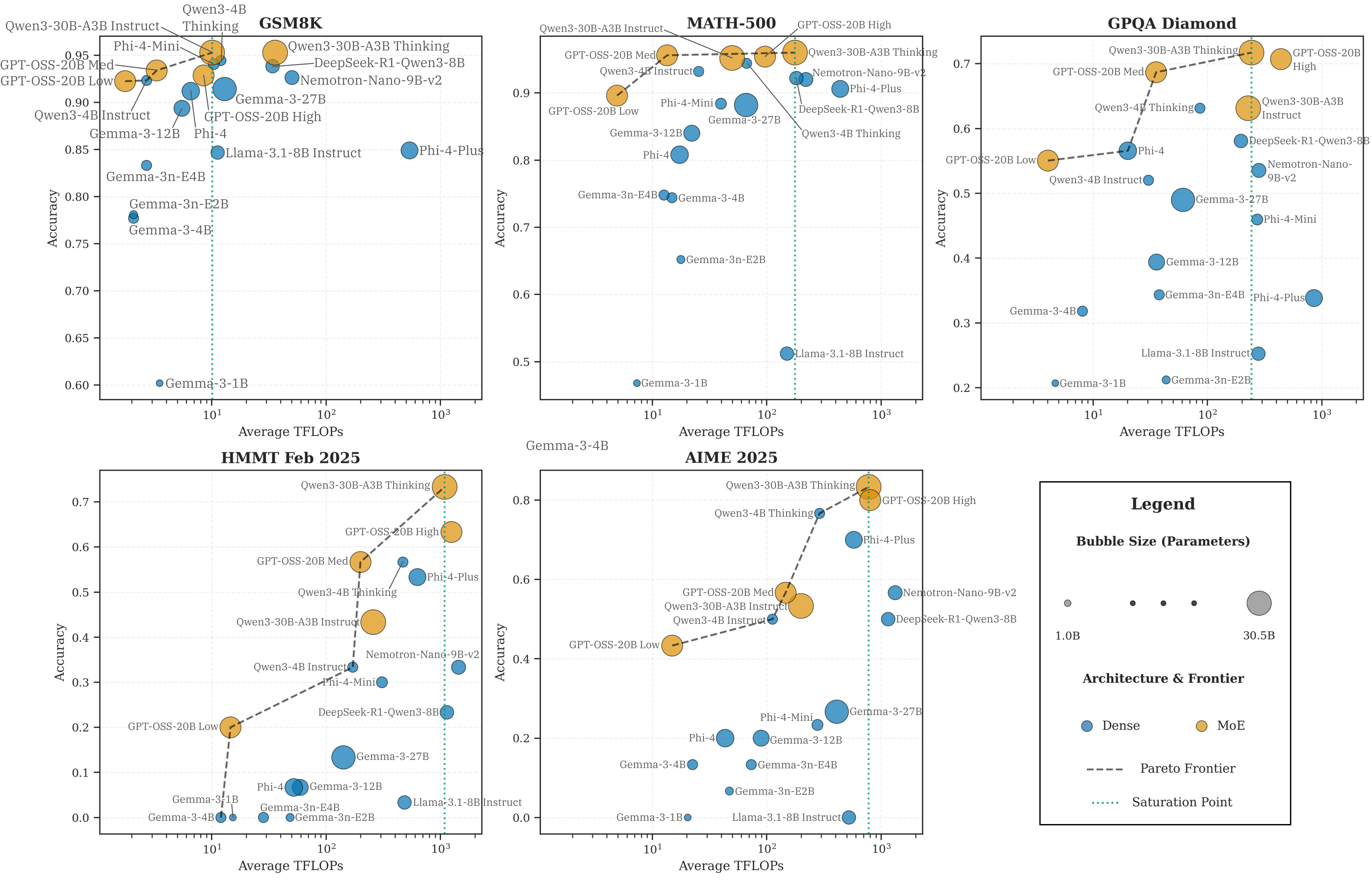}
    \caption{\textbf{The Reasoning Pareto Frontier.}
    Each point represents the efficiency of a model on the given benchmark. The $x$-axis denotes the average FLOP estimation per query (log scale), while the $y$-axis tracks task accuracy. \textbf{Color} distinguishes model architecture (Dense: blue, MoE: yellow), and \textbf{marker size} is proportional to parameter count. Note the task-dependent saturation points, particularly visible in the lighter GSM8K versus the steeper ascent in the reasoning heavy AIME-2025 benchmark.}
    \label{fig:pareto-detailed}
\end{figure*}

The landscape of reasoning efficiency, visualized in Figure~\ref{fig:pareto-detailed}, reveals a consistent piecewise-linear relationship between inference compute and accuracy across varying domains. In the initial "linear" regime, exponential increases in floating-point operations yield rapid improvements in task performance; however, this is invariably followed by a task-dependent saturation point. The location of the saturation 'knee' serves as a proxy for the alignment between model capability and task difficulty. On lower-complexity benchmarks like \textsc{gsm8k} and \textsc{MATH500}, models possess high intrinsic proficiency, causing them to hit the reasoning ceiling rapidly at moderate compute budgets ($<10^{16}$ FLOPs). Conversely, for complex frontiers such as \textsc{AIME25} and \textsc{GPQA}, this saturation point is shifted to significantly higher compute regimes. However, the flattening of the efficiency curve even in these steep regimes underscores a limitation. While additional inference compute unlocks performance up to a point, it cannot bridge the gap when a problem’s complexity exceeds the model’s fundamental deductive horizon.

\subsection{Model Parameter saturates against inference time compute.} 
\label{sec:parameter-saturation}
Within the landscape of Figure~\ref{fig:pareto-detailed}, we observe a distinct \emph{model parameter saturation} effect among \texttt{dense} architectures. Raw parameter scaling yields diminishing returns relative to reasoning length. Specifically, smaller models (1B--8B range) demonstrate a remarkable capability to challenge significantly larger counterparts by leveraging extended reasoning sequences. These models effectively substitute inference-time compute for parameter capacity, often achieving accuracy parity with 30B+ baselines while incurring a \emph{lower compute cost}. This empirically validates the that "thinking compute" acts as a fungible resource, capable of efficiently compensating for limited architectural capacity.

\subsection{Architectural Advantage of MoE.}
\label{sec:architecture-analysis}
Our evaluation isolates a distinct structural advantage for sparse architectures in reasoning-intensive tasks. As illustrated by the yellow markers in Figure~\ref{fig:pareto-detailed}, MoE models systematically dominate the Pareto frontier, particularly on benchmarks requiring extended reasoning like GPQA-Diamond, AIME 2025 and HMMT Feb 2025. By activating a sparse subset of parameters per token, MoE architectures decouple generation length from total FLOP consumption. This allows them to generate significantly longer reasoning chains while remaining within the same computational budget. Consequently, sparsity appears to be a critical enabler for CoT scaling, offering a better conversion rate of FLOPs to reasoning accuracy.

\subsection{Temporal Trends}
\label{sec:temporal-trends}
To quantify the rate of progress in reasoning capability relative to computational cost, we define a scalar \textbf{Efficiency Score} ($S_{\text{eff}}$) for each model. This metric represents the ratio between a model's mean performance and its average computational expenditure per query:

\begin{equation}
    S_{\text{eff}} = \frac{\frac{1}{N}\sum_{i=1}^{N} \mathcal{A}_i}{\log_{10}\left(\frac{1}{N}\sum_{i=1}^{N} \text{FLOPs}_i\right)}
\end{equation}

where $N=5$ is the number of benchmarks, $\mathcal{A}_i$ is the accuracy on benchmark $i$, and $\text{FLOPs}_i$ is the average floating-point operations consumed to generate a single reasoning trace for that benchmark. By utilizing a logarithmic denominator, $S_{\text{eff}}$ effectively measures the \emph{accuracy return per order of magnitude of compute}, penalizing models that achieve marginal gains through exponential increases in cost while rewarding efficient architectures and concise reasoning traces.

\begin{figure}[ht]
    \centering
    \includegraphics[width=\linewidth]{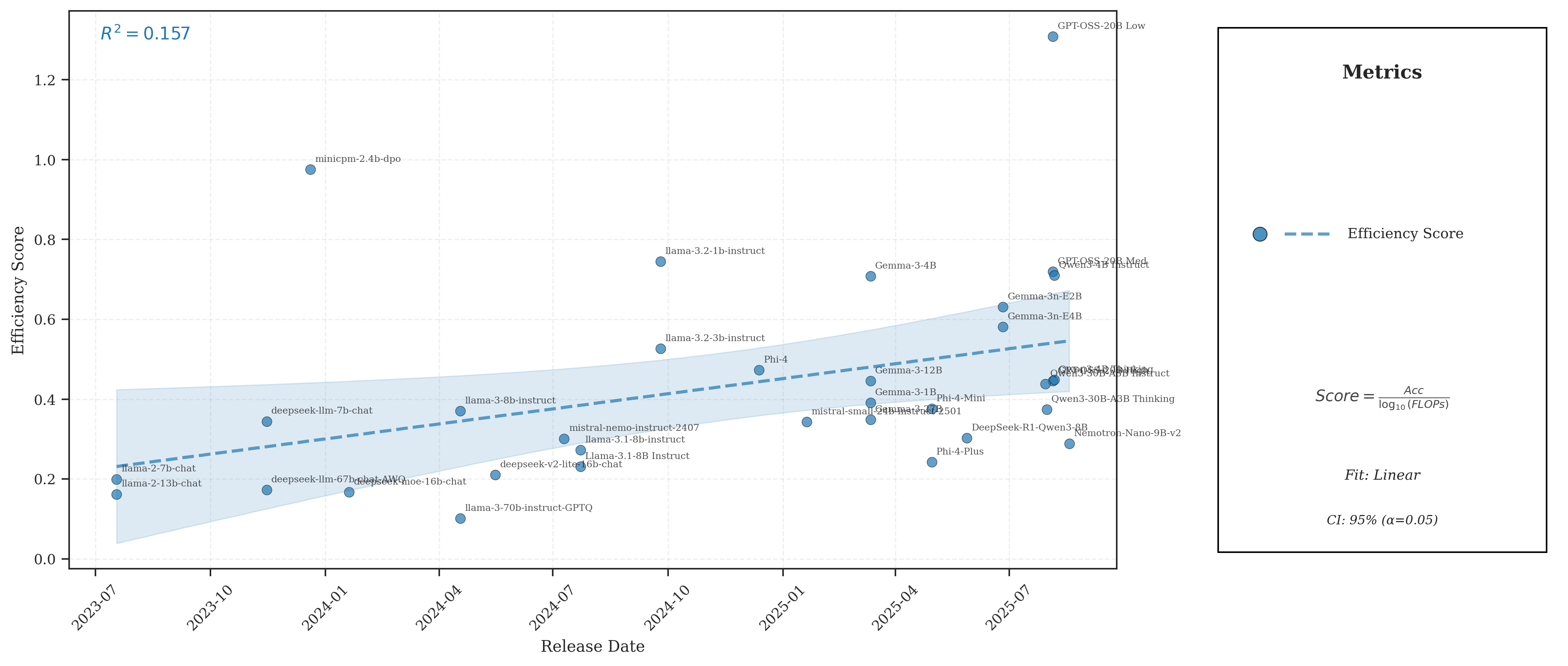}
    \caption{\textbf{Temporal Progress.}
    Efficiency score (average accuracy across all five benchmarks normalized by $\log_{10}(\text{FLOPs})$) versus model release date. The trend indicates a consistent improvement in reasoning efficiency, with increasing variance in 2025 driven by specialized reasoning models.}
    \label{fig:temporal}
\end{figure}

As seen on Figure~\ref{fig:temporal}, aggregating these metrics chronologically indicates a general upward trajectory in $S_{\text{eff}}$ from mid-2023 through mid-2025. Notably, the variance in this metric has expanded over time, signaling a divergence between general-purpose models and specialized reasoning variants. This trend suggests that recent algorithmic innovations, specifically reasoning-focused fine-tuning and reinforcement learning strategies are successfully shifting the Pareto frontier outward, enabling newer models to solve complex problems with less computational overhead than their predecessors.

\paragraph{Limitations.} However, this linear trend requires careful interpretation. There is no trivial, monotonic function linking reasoning compute to accuracy; as discussed in Section~\ref{sec:main-pareto}, the true relationship is characterized by task-dependent saturation points rather than infinite growth. Furthermore, this efficiency metric is intrinsically bound to the specific difficulty distribution of our evaluation suite. While the upward trajectory indicates genuine progress in algorithmic efficiency, it captures performance only within the context of current academic benchmarks, not necessarily generalizing to open-ended or undefined problem domains.

\subsection{Computational Cost of Failure}

\begin{figure}[t]
    \centering
    \includegraphics[width=0.95\linewidth]{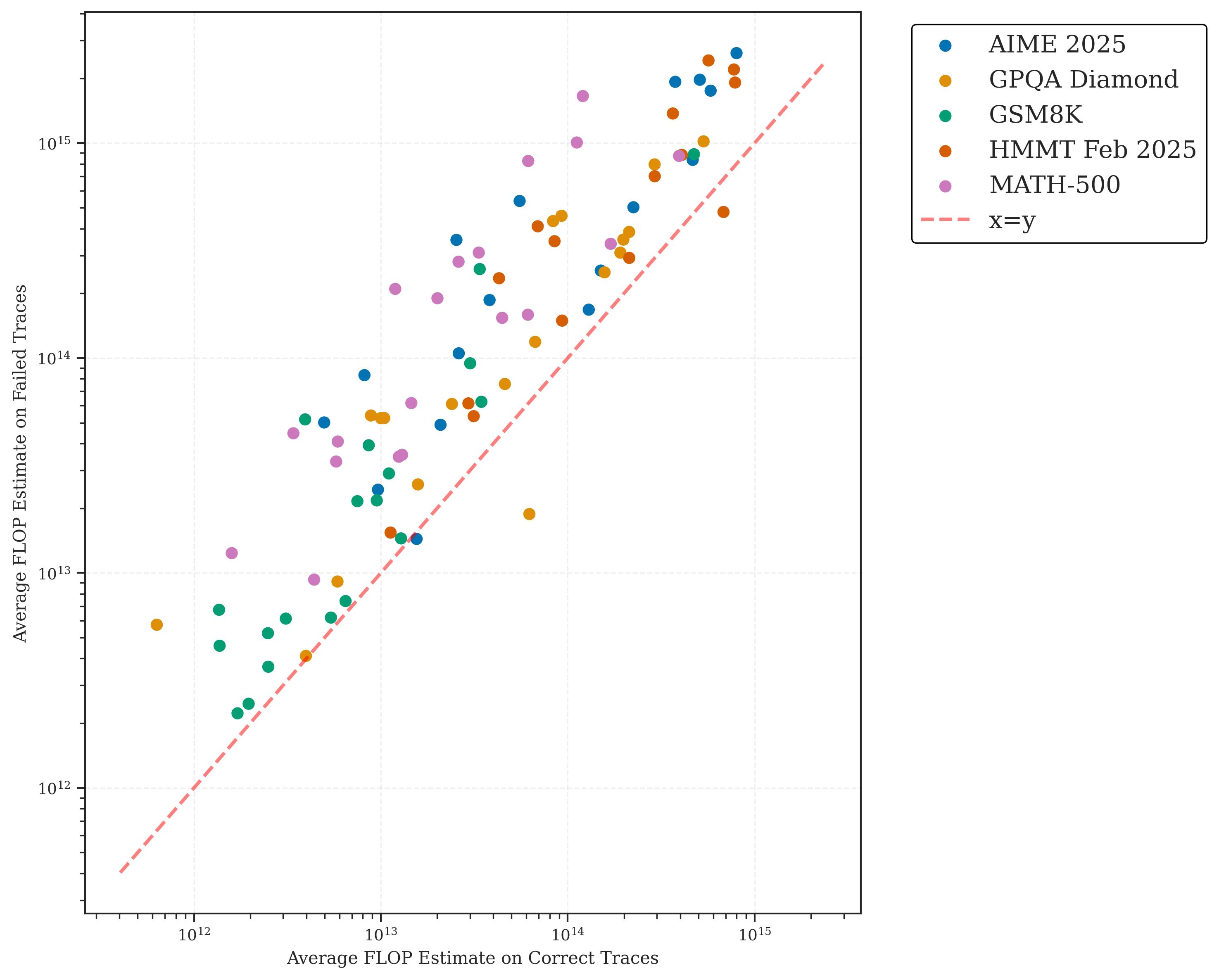}
    \caption{\textbf{Trace Length Asymmetry.}
    Average compute cost of \emph{incorrect} versus \emph{correct} traces. Points above the red dashed line ($y=x$) indicate models that spend more compute when failing. 97\% of evaluated models exhibit this inefficiency.}
    \label{fig:trace-length}
    \vspace{-10pt} 
\end{figure}

Shown on Figure~\ref{fig:trace-length}, our empirical analysis finds that 97\% of the evaluated models expend significantly more compute on incorrect reasoning traces than on correct derivations, thereby further reinforcing the known trace length asymmetry of reasoning models. While the \emph{deductive horizon} discussed in Section~\ref{sec:main-pareto} establishes a hard ceiling on accuracy, this asymmetry demonstrates that exceeding intrinsic model capabilities incurs a disproportionate computational penalty.  Rather than terminating generation under high uncertainty, models frequently engage in prolonged hallucinations, recursive looping, or divergence into flawed logical paths. This results in compute burn where the inference engine exhausts the available context window in a futile attempt to resolve ambiguity. Consequently, optimizing test-time compute presents a significant algorithmic challenge: distinguishing between the extended duration required for valid complex deduction and the wasteful latency of failure without inadvertently truncating necessary reasoning steps remains non-trivial. \textbf{Although our study is delimited to open-weights for reproducibility, we hypothesize that this asymmetry is not an artifact of specific architectures but an intrinsic characteristic of autoregressive reasoning, likely extending to closed-source frontier models as well.}

\section{Conclusion}

In this work, we presented a comprehensive, compute-aware evaluation of reasoning models, establishing the empirical Pareto frontiers between inference cost and task accuracy. Our analysis yields three critical insights for the development and deployment of reasoning systems.

First, we identify a distinct architectural advantage for \textbf{Mixture-of-Experts (MoE) }models. By decoupling parameter count from active FLOPs, MoEs consistently dominate the Pareto frontier, achieving higher accuracy per unit of compute compared to dense counterparts. This suggests that sparsity is a fundamental enabler for efficient test-time scaling.

Second, we uncover the phenomenon of \textbf{Trace Length Asymmetry}, where incorrect reasoning traces are systematically longer than correct ones. This contradicts the assumption that extended computation monotonically improves performance; rather, ungrounded reasoning often leads to verbose failure modes.

Finally, we demonstrate that inference-time scaling exhibits clear \textbf{saturation points}. While Chain-of-Thought enables smaller models to punch above their weight, there exists a task-dependent ``knee'' in the Pareto curve beyond which additional FLOPs provide negligible returns. Future work in reasoning efficiency must therefore focus not only on expanding token budgets but on intelligent, early-stopping mechanisms that recognize when a model has exhausted its reasoning capabilities.

\paragraph{Broader Impacts.}
As Large Language Models transition from static knowledge retrievers to active reasoning agents, the computational burden shifts from pre-training to inference. Our characterization of the Pareto frontier offers a roadmap for mitigating the environmental impact of this shift; by identifying architectures that maximize reasoning yield per FLOP, we provide actionable insights for reducing the energy consumption and carbon footprint of large-scale deployments. Furthermore, establishing clear efficiency metrics democratizes access to advanced capabilities, allowing resource-constrained entities to select models that balance performance with economic viability.

\bibliographystyle{plainnat}
\bibliography{refs}

\section{Appendix}

\paragraph{Architectural Configurations.}
Table~\ref{tab:model-configs} summarizes representative model configurations. Note the significant variation in the expansion factor $d_{\text{ff}}/d$: ranging from 1.00 (gpt-oss-20b-reason) to 8.00 (gemma-3n-E4B-it), which directly affects the FFN share of total compute and memory bandwidth requirements.

\begin{table}[ht]
\centering
\caption{Architectural parameters for provided model configurations.}
\label{tab:model-configs}
\scriptsize
\begin{tabular}{@{}lcccccccccccc@{}}
\toprule
\textbf{Model} & $L$ & $d$ & $d_{\text{ff}}$ & $d_{\text{ff}}/d$ & $V$ & $n_h$ & $n_{kv}$ & \textbf{Gated} & $E$ & $k$ \\
\midrule
Qwen3-30B-A3B-Thinking-2507 & 48 & 2048 & 6144 & 3.00 & 151936 & 32 & 4 & true & 128 & 8 \\
Qwen3-30B-A3B-Instruct-2507 & 48 & 2048 & 6144 & 3.00 & 151936 & 32 & 4 & true & 128 & 8 \\
Qwen3-4B-Thinking-2507 & 36 & 2560 & 9728 & 3.80 & 151936 & 32 & 8 & true & 0 & 0 \\
Qwen3-4B-Instruct-2507 & 36 & 2560 & 9728 & 3.80 & 151936 & 32 & 8 & true & 0 & 0 \\
gpt-oss-20b-reason & 24 & 2880 & 2880 & 1.00 & 200000 & 64 & 8 & true & 32 & 4 \\
Nemotron-Nano-9B-v2 & 56 & 4480 & 15680 & 3.50 & 131072 & 40 & 8 & false & 0 & 0 \\
DeepSeek-R1-0528-Qwen3-8B & 36 & 4096 & 12288 & 3.00 & 151936 & 32 & 8 & true & 0 & 0 \\
Llama-3.1-8B-Instruct & 32 & 4096 & 14336 & 3.50 & 128256 & 32 & 8 & true & 0 & 0 \\
gemma-3-1b-it & 26 & 1152 & 6912 & 6.00 & 262144 & 4 & 1 & true & 0 & 0 \\
gemma-3-4b-it & 34 & 2560 & 10240 & 4.00 & 256000 & 8 & 2 & true & 0 & 0 \\
gemma-3-12b-it & 48 & 3840 & 15360 & 4.00 & 256000 & 16 & 8 & true & 0 & 0 \\
gemma-3-27b-it & 62 & 5376 & 21504 & 4.00 & 256000 & 32 & 16 & true & 0 & 0 \\
gemma-3n-E2B-it & 30 & 2048 & 8192 & 4.00 & 262400 & 8 & 2 & true & 0 & 0 \\
gemma-3n-E4B-it & 35 & 2048 & 16384 & 8.00 & 262400 & 8 & 2 & true & 0 & 0 \\
phi-4 & 40 & 5120 & 17920 & 3.50 & 100352 & 40 & 10 & false & 0 & 0 \\
Phi-4-mini-reasoning & 32 & 3072 & 8192 & 2.67 & 200064 & 24 & 8 & false & 0 & 0 \\
Phi-4-reasoning-plus & 40 & 5120 & 17920 & 3.50 & 100352 & 40 & 10 & false & 0 & 0 \\
\bottomrule
\end{tabular}
\end{table}

\begin{figure}[th]
    \centering
    \includegraphics[width=\textwidth]{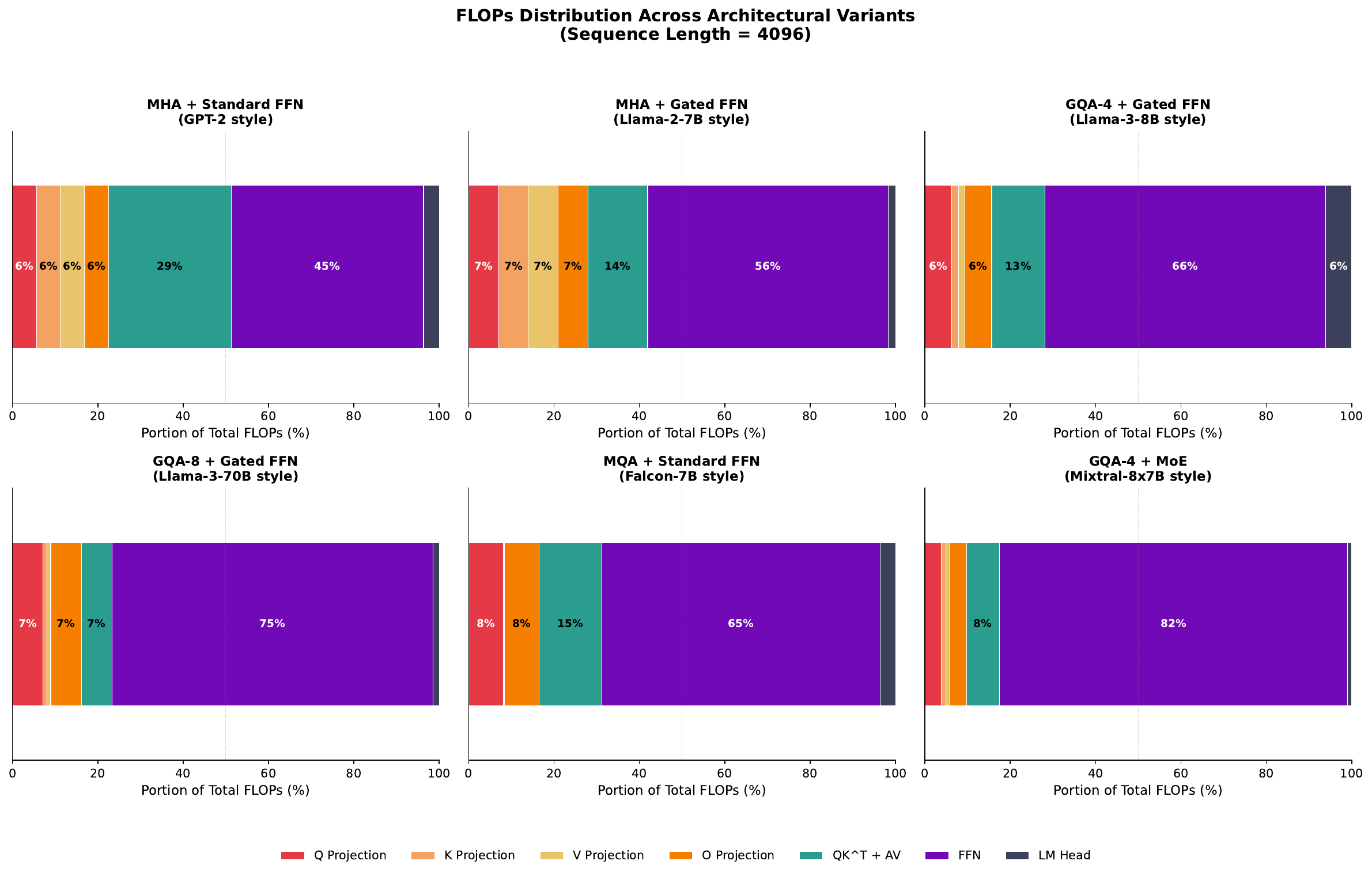}
    \caption{FLOPs distribution across architectural variants at sequence length 4096. GQA reduces K/V projection costs (orange/yellow), while gated FFN (SwiGLU) increases the FFN share (purple). MoE models show dominant FFN contribution due to sparse expert activation.}
    \label{fig:flops-breakdown}
\end{figure}

\paragraph{Compute Distribution.}
Figure~\ref{fig:flops-breakdown} illustrates how architectural choices affect FLOPs distribution at sequence length 4096. Three patterns emerge:
\begin{enumerate}
    \item \textit{FFN dominates}: The feed-forward network accounts for 50--75\% of FLOPs across dense models, rising to 82\% for MoE due to sparse expert activation.
    \item \textit{GQA reduces projections}: Moving from MHA to GQA-8 shrinks attention projection costs from 28\% to 16\% of total FLOPs.
    \item \textit{Quadratic attention is small at moderate lengths}: At 4K tokens, attention compute (QK$^\top$ + AV) contributes only 7--15\% of FLOPs.
\end{enumerate}

\begin{figure}[H]
    \centering
    \includegraphics[width=\textwidth]{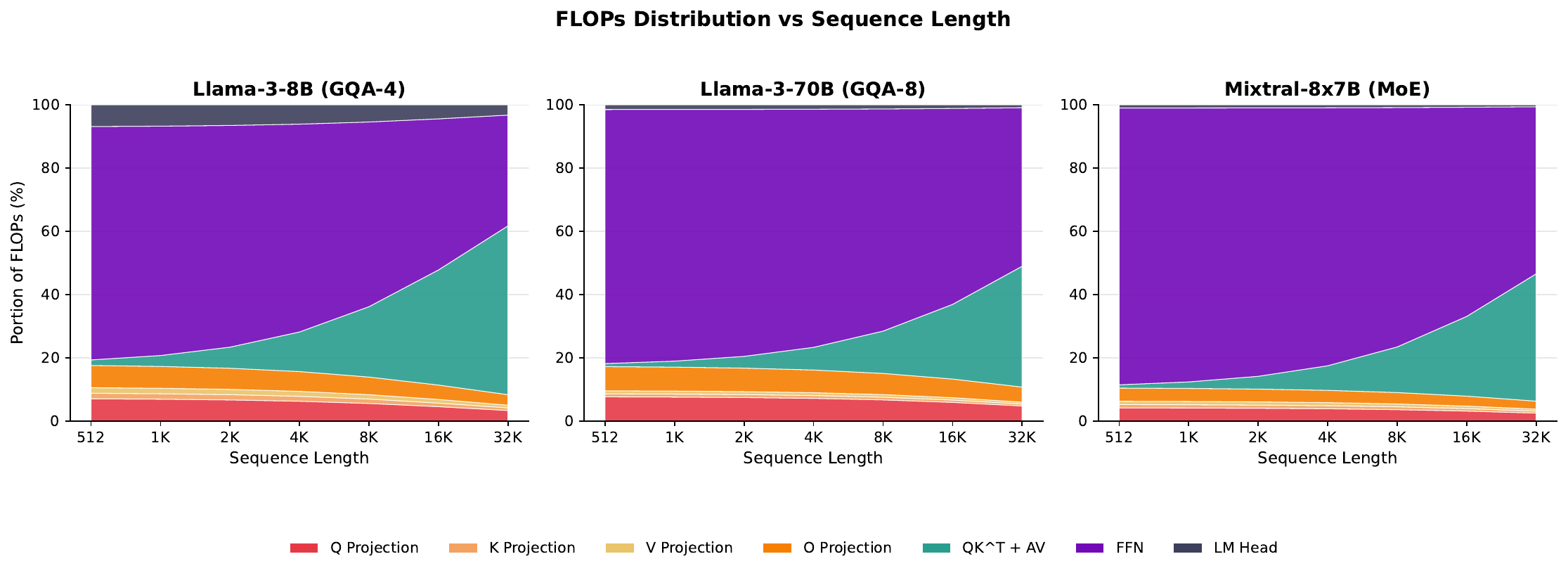}
    \caption{FLOPs distribution versus sequence length for Llama-3-8B (GQA-4), Llama-3-70B (GQA-8), and Mixtral-8$\times$7B (MoE). The quadratic attention component (teal) grows from under 5\% at short sequences to over 50\% at 32K tokens.}
    \label{fig:flops-scaling}
\end{figure}

\paragraph{Sequence Length Scaling.}
The quadratic attention term grows with sequence length. Figure~\ref{fig:flops-scaling} shows this scaling for three representative models: attention compute rises from ${\sim}2\%$ at 512 tokens to ${\sim}12\%$ at 4K, ${\sim}35\%$ at 16K, and over 50\% at 32K tokens. This crossover motivates efficient attention mechanisms for long-context inference.

\paragraph{Relation to $\mathbf{2N}$ Approximation.}
When sequence length is small relative to model width ($s \ll d$), the quadratic term becomes negligible and Equation~\ref{eq:flops-total} reduces to approximately $2N$ FLOPs per token. \citet{kaplan2020scaling} showed that for GPT-3 ($d = 12288$, $s = 4096$), the quadratic term contributes less than 3\% of total FLOPs. However, for long-context reasoning with $s > 8K$, the full expression is necessary.

\paragraph{Comparison with Published Results.}
To contextualize our evaluation pipeline, Table \ref{tab:published-results} compares the officially reported accuracies of each model with the results we obtained under our unified benchmarking environment, which uses AA–style prompting, generation temperatures and a fixed single-pass CoT protocol. This side-by-side view highlights how models behave under consistent inference settings, revealing where our measurements closely track the published numbers and where deviations arise due to differences in decoding parameters, prompt formats, or majority-vote aggregation commonly used in official reports.

\begin{table}[H]
\centering
\caption{Published results for each evaluated model across the AIME25, HMMT Feb 25, GPQA-Diamond, MATH-500 and GSM8k benchmarks. A dash (--) indicates that no official number was available at the time of writing. For each cell, the first value is the officially reported accuracy and the second value is our reproduced accuracy.}
\label{tab:published-results}
\scriptsize
\begin{tabular}{@{}lccccc@{}}
\toprule
\textbf{Model} & \textbf{AIME25} & \textbf{HMMT Feb 25} & \textbf{GPQA-Diamond} & \textbf{MATH-500} & \textbf{GSM8k} \\
\midrule
Qwen3-30B-A3B-Thinking-2507 & 85.0 / 83.3 & 71.4 / 73.3 & 73.4 / 71.7 & -- / 96.0 & -- / 95.3 \\
Qwen3-30B-A3B-Instruct-2507 & 61.3 / 53.3 & 43.0 / 43.3 & 70.4 / 63.1 & -- / 95.2 & -- / 95.3 \\
Qwen3-4B-Thinking-2507 & 81.3 / 76.7 & 55.5 / 56.7 & 65.8 / 63.1 & -- / 94.4 & -- / 94.5 \\
Qwen3-4B-Instruct-2507 & 47.4 / 50.0 & 31.0 / 33.3 & 62.0 / 52.0 & -- / 93.2 & -- / 92.3 \\
gpt-oss-20b-reason-low & 37.1 / 43.3 & -- / 2. & 56.8 / 55.0 & -- / 89.6 & -- / 92.2 \\
gpt-oss-20b-reason-medium & 72.1 / 56.6 & -- / 56.6 & 66.0 / 68.6 & -- / 95.6 & -- / 93.4 \\
gpt-oss-20b-reason-high & 91.7 / 80.0 & -- / 63.3 & 71.5 / 70.7 & -- / 95.4 & -- / 92.9 \\
Nemotron-Nano-9B-v2 & 72.1 / 56.7 & -- / 33.3 & 64.0 / 53.5 & 97.8 / 92.0 & -- / 92.6 \\
DeepSeek-R1-0528-Qwen3-8B & 76.3 / 50.0 & 61.5 / 23.3 & 61.1 / 58.1 & -- / 92.2 & -- / 93.9 \\
Llama-3.1-8B-Instruct & -- / 0.0 & -- / 3.3 & -- / 25.3 & -- / 51.2 & 84.5 / 84.7 \\
gemma-3-1b-it & -- / 0.0 & -- / 0.0 & 15.0 / 20.7 & 24.2 / 46.8 & 38.4 / 60.2 \\
gemma-3-4b-it & -- / 13.3 & -- / 0.0 & 25.4 / 31.8 & 43.3 / 74.4 & 71.0 / 77.7 \\
gemma-3-12b-it & -- / 20.0 & -- / 6.7 & 24.3 / 39.4 & 50.0 / 84.0 & 82.6 / 89.4 \\
gemma-3-27b-it & -- / 26.7 & -- / 13.3 & 42.4 / 49.0 & 69.0 / 88.2 & 95.9 / 91.4 \\
gemma-3n-E2B-it & -- / 6.7 & -- / 0.0 & -- / 21.2 & -- / 65.2 & -- / 78.1 \\
gemma-3n-E4B-it & -- / 13.3 & -- / 0.0 & -- / 34.3 & -- / 74.8 & -- / 83.3 \\
phi-4 & 62.9 / 20.0 & -- / 6.7 & 65.8 / 56.6 & -- / 80.8 & -- / 91.2 \\
Phi-4-mini-reasoning & 31.77 / 23.3 & -- / 30.0 & 44.51 / 46.0 & 91.20 / 88.4 & -- / 94.1 \\
Phi-4-mini-flash-reasoning & 33.59 / -- & -- / -- & 45.08 / -- & 92.45 / -- & -- / -- \\
Phi-4-reasoning-plus & 78.0 / 70.0 & -- / 53.3 & 68.9 / 33.8 & -- / 90.6 & -- / 84.9 \\
\bottomrule
\end{tabular}
\end{table}
\end{document}